\documentclass[11pt, a4paper, logo, copyright]{deepmind}
\pdftrailerid{redacted}

\makeatletter
\renewcommand\bibentry[1]{\nocite{#1}{\frenchspacing\@nameuse{BR@r@#1\@extra@b@citeb}}}
\makeatother

\usepackage{kantlipsum, lipsum}
\usepackage{dsfont}
\usepackage{dm-colors}

\usepackage[authoryear, sort&compress, round]{natbib}

\graphicspath{{figures/}}

\title{PGMax: Factor Graphs for Discrete Probabilistic Graphical Models and Loopy Belief Propagation in JAX}

\correspondingauthor{stannis@deepmind.com, adedieu@deepmind.com}


\reportnumber{} 

\renewcommand{\today}{2023-02-02}

\author[1]{Guangyao Zhou}
\author[1]{Antoine Dedieu}
\author[2]{Nishanth Kumar}
\author[1]{Wolfgang Lehrach}
\author[1]{Miguel L\'azaro-Gredilla}
\author[1]{Shrinu Kushagra}
\author[1]{Dileep George}

\affil[1]{DeepMind}
\affil[2]{Massachusetts Institute of Technology}

\begin{abstract}
	PGMax is an open-source Python package for (a) easily specifying discrete Probabilistic Graphical Models (PGMs) as factor graphs; and (b) automatically running efficient and scalable loopy belief propagation (LBP) in JAX. PGMax supports general factor graphs with tractable factors, and leverages modern accelerators like GPUs for inference. Compared with existing alternatives, PGMax obtains higher-quality inference results with up to three orders-of-magnitude inference time speedups. PGMax additionally interacts seamlessly with the rapidly growing JAX ecosystem, opening up new research possibilities. Our source code, examples and documentation are available at \url{https://github.com/deepmind/PGMax}.
\end{abstract}

\begin{document}

\maketitle

\section{Introduction}

Discrete Probabilistic Graphical Models (PGMs) compactly encode the full joint probability distribution of a set of discrete variables. Discrete PGMs are commonly specified using factor graphs~\citep{kschischang2001factor}, and have seen successful applications in a wide range of fields such as computer vision~\citep{wang2013markov}, natural language processing~\citep{wang2019bert} and biology~\citep{mora2011biological}. A standard approach to run inference on discrete PGMs with tractable factors involves message-passing algorithms like Loopy Belief Propagation (LBP)~\citep{pearl1988probabilistic, murphy1999loopy}. Given a factor graph with discrete variables and tractable factors, LBP propagates easy-to-derive message updates between the variables and the factors of the graph. However, despite the many past attempts at implementing factor graphs and LBP in Python\footnote{Some examples include \href{https://github.com/rdlester/pyfac}{\texttt{pyfac}}, \href{https://github.com/mbforbes/py-factorgraph}{\texttt{py-factorgraph}}, \href{https://github.com/jeroen-chua/factorflow}{\texttt{factorflow}}, \href{https://github.com/danbar/fglib}{\texttt{fglib}}, \href{https://github.com/ilyakava/sumproduct}{\texttt{sumproduct}}, \href{https://github.com/pgmpy/pgmpy}{\texttt{pgmpy}}, \href{https://github.com/jmschrei/pomegranate}{\texttt{pomegranate}}.}, there is no well-established open-source Python package that implements efficient and scalable LBP for general factor graphs. In particular, a key challenge lies in the design and manipulation of the Python data structures containing the LBP messages for factor graphs with (a) complicated topology; (b) complex factor definitions; and (c) discrete variables with a varying number of states.

In this paper, we describe PGMax, a new open-source Python package that (a) provides an easy-to-use interface for specifying general factor graphs with discrete variables and tractable factors and (b) implements efficient and scalable LBP in JAX~\citep{jax2018github}. Compared with existing alternative Python packages, PGMax (a) supports a larger class of tractable factors; (b) demonstrates superior inference performance in terms of quality, inference speed and scalability; and (c) opens up new research possibilities from its seamless interaction with the rapidly growing JAX ecosystem~\citep{deepmind2020jax}.

\section{Related work}

Several Python packages have been proposed for running LBP in discrete PGMs: two of the most recent ones are \texttt{pomegranate}~\citep{schreiber2017pomegranate} and \texttt{pgmpy}~\citep{ankan2015pgmpy}. However, these two packages suffer from severe limitations. First, they only support a limited set of discrete factors (e.g, they do not support the logical factors as we do---see Section \ref{sec:features}). Second, their inference implementations make heavy use of control flows with irregular Python data structures, and the lack of efficient array-based implementations largely restricts their computations to CPUs. As we show in Sec.~\ref{sec:comparison}, these limitations severely affect the packages' efficiency and scalability.

Due to the challenges in deriving an efficient implementation of LBP in Python, past works resort to C++ or Julia. \texttt{OpenGM}~\citep{andres2012opengm, opengm-benchmark} supports general factor graphs but is no longer maintained, while \texttt{ForneyLab}~\citep{ForneyLab.jl-2019} and \texttt{ReactiveMP}~\citep{bagaev2021reactivemp} focus mostly on conjugate state-space models. However, these packages cannot interact with the vast Python scientific computing ecosystem.

Another related line of work is probabilistic programming languages (PPLs). While PPLs~\citep{van2018introduction} ~\citep{Carpenter2017-zi, bingham2018pyro, Salvatier2016, phan2019composable, ge2018t} have appealing properties, they typically focus on PGMs with continuous variables, with Markov Chain Monte Carlo methods like Hamiltonian Monte Carlo as the core inference engine, and have limited support for undirected PGMs and discrete variables~\citep{zhou2019mixed}.

In contrast with these existing methods, the package we introduce in this paper, PGMax, is a specialized PPL that provides an interface for specifying discrete PGMs with tractable factors; and that automatically derives GPU-accelerated inference via LBP.

\section{Package features}\label{sec:features}

We describe below some of the appealing features of PGMax.

\textbf{Easy specification of general factor graphs: } PGMax handles discrete variables with a varying number of states. It also allows to easily specify general factor graphs with complicated topology and with complex factor definitions---as we discuss in Sec.~\ref{sec:fg}.

\noindent\textbf{Efficient, scalable LBP implementation: } PGMax adopts an LBP implementation using parallel message updates and damping~\citep{pretti2005message}. This setup has been extensively tested in recent works~\citep{dedieu2023learning, george2017generative, lazaro2021query, lazaro2021perturb, zhou2021graphical} on a wide range of discrete PGMs. To this end, PGMax develops a novel fully flat array-based LBP implementation (see Sec.~\ref{sec:implementation}) in JAX, which makes it able to effectively leverage just-in-time compilation and to run on modern accelerators like GPUs. Consequently, as we show in Sec.~\ref{sec:comparison}, the resulting inference is up to three orders of magnitude faster than existing alternatives.

\noindent\textbf{Logical factors with linear message updates: } Logical factors can be used in discrete PGMs to represent logical relations between variables \citep{ravanbakhsh2016boolean,lazaro2021perturb,dedieu2023learning}. PGMax supports three types of \href{https://github.com/deepmind/PGMax/blob/main/pgmax/factor/logical.py}{logical factors}. For each type, it derives specialized efficient message updates with optimized complexity linear in the number of variables connected to the logical factors---which allows PGMax to support large logical factors.

\noindent\textbf{Seamless interaction with JAX: } PGMax implements LBP as pure functions with no side effects. This functional design allows PGMax to seamlessly interact with the rapidly growing JAX ecosystem~\citep{deepmind2020jax}, and opens up exciting new possibilities---as we discuss in Sec.~\ref{sec:jax}.

\noindent\textbf{Software engineering best practices: } PGMax follows professional software development workflows, with enforced format and static type checking, automated continuous integration and documentation generation, and comprehensive unit tests that fully cover the codebase.

\section{A fully flat array-based LBP implementation}\label{sec:implementation}

LBP passes real-valued vectors, called ``messages'', along the edges of a factor graph. For a given factor graph, the message updates can be easily derived. However, it is empirically challenging to design efficient Python data structures for storing and updating the messages due to the (a) potentially complicated topology of the factor graph (b) complex factor definitions and (c) varying number of states of the discrete variables. Existing Python packages (e.g. \texttt{pgmpy} and \texttt{pomegranate}) typically resort to irregular data structures to store the messages and heavily use control flows for message updates. This makes them inefficient, and largely restricts their computations to CPUs. A possible alternative here is to pad messages into regular multidimensional arrays and update messages using efficient array-based computations. However, this would require model-specific data structures and padding strategies, and we have empirically observed that this leads to inefficient memory usage.

To address these challenges, PGMax concatenates all the variable-to-factor and factor-to-variable messages of a factor graph into two separate 1D arrays, and develops a novel fully flat array-based LBP implementation. By keeping track of the global indices of variable states and the corresponding valid factor configurations, PGMax leverages the \href{https://www.tensorflow.org/xla/operation_semantics#gather}{gather}/\href{https://www.tensorflow.org/xla/operation_semantics#scatter}{scatter} operations \href{https://jax.readthedocs.io/en/latest/jax.ops.html#indexed-update-operators}{implemented in JAX} for a succinct, efficient and scalable LBP implementation.

\section{Experiments}\label{sec:experiments}
We refer the readers to our \href{https://github.com/deepmind/PGMax/tree/main/examples}{example notebooks} for detailed tutorials on basic PGMax usage.

\subsection{Comparison with existing alternatives}\label{sec:comparison}

We compare PGMax with existing Python alternatives for maximum-a-posteriori (MAP) inference on randomly generated restricted Boltzmann Machines (RBMs). The exisiting packages we consider are \href{https://github.com/mbforbes/py-factorgraph}{\texttt{py-factorgraph}}, \href{https://github.com/danbar/fglib}{\texttt{fglib}}, \href{https://github.com/pgmpy/pgmpy}{\texttt{pgmpy}} and \href{https://github.com/jmschrei/pomegranate}{\texttt{pomegranate}}. \texttt{py-factorgraph} and \texttt{fglib} are non-functional. \texttt{pgmpy}'s \href{https://pgmpy.org/exact_infer/bp.html}{belief propagation} uses junction tree algorithm for exact inference; however it cannot be used even for the 30-unit RBM considered here due to slow inference speed. We then exclude \texttt{py-factorgraph}, \texttt{fglib} and \texttt{pgmpy}'s BP from comparison. 
\texttt{pgmpy} additionally implements the \href{https://pgmpy.org/exact_infer/mplp.html}{max-product linear programming (MPLP)} algorithm~\citep{globerson2007fixing}, which we use as a baseline, along with \texttt{pomegranate}'s \href{https://pomegranate.readthedocs.io/en/latest/MarkovNetwork.html#pomegranate.MarkovNetwork.MarkovNetwork.predict}{LBP}. For \texttt{pgmpy}'s MPLP and \texttt{pomegranate}'s LBP, we use the default settings. For PGMax, we use $200$ LBP iterations and a $0.5$ damping. For each method, we measure its inference quality by computing the potential of the MAP configuration returned.

\texttt{pgmpy} only runs on CPUs. Although \texttt{pomegranate} supports CPU multiprocessing and GPUs, we did not see any obvious speedups from these advanced features. Consequently, we only use CPUs for both \texttt{pgmpy} and \texttt{pomegranate}. We test PGMax on both CPUs (using Intel(R) Core(TM) i7-7820HQ) and GPUs (using NVIDIA Quadro M1200).

\begin{figure}[ht!]
	\centering
	\captionsetup{justification=centering}
	\includegraphics[width=0.75\textwidth]{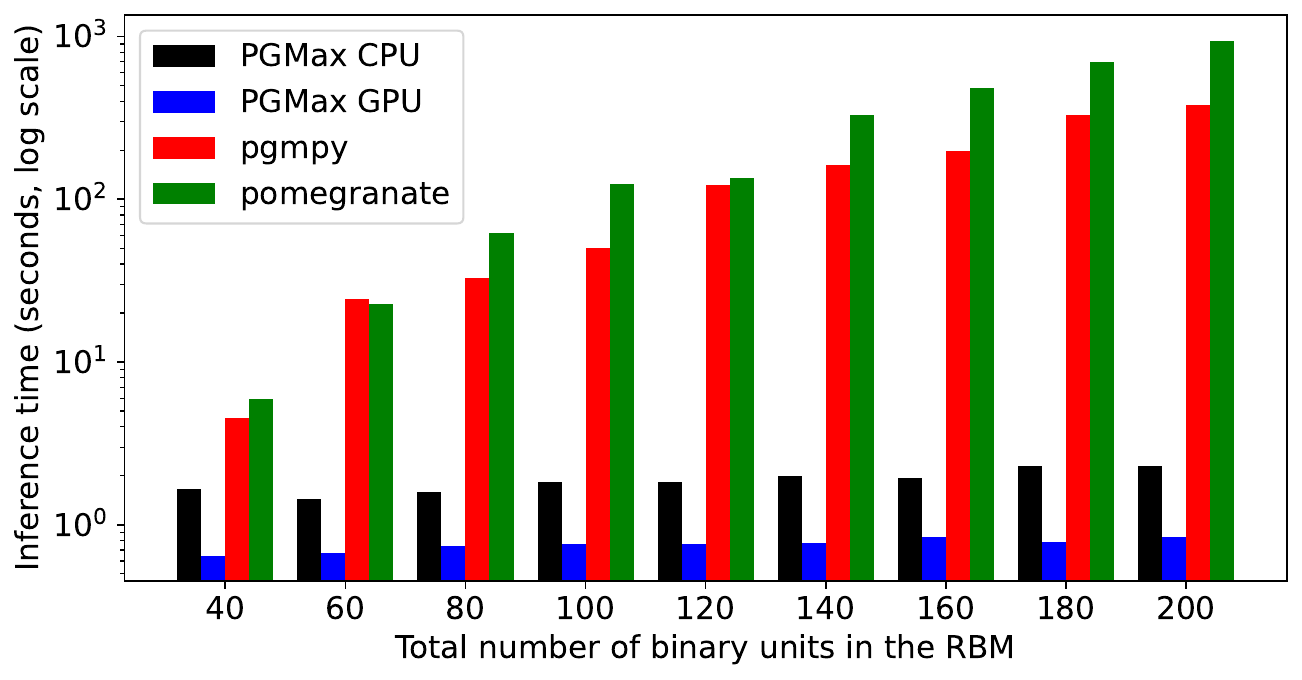}
	\caption{MAP inference time of the methods compared on RBMs with increasing sizes. PGMax on a GPU is three orders of magnitude faster than existing alternatives.}
	\label{fig:results}
\end{figure}
Our first experiment considers $20$ randomly generated RBMs with $30$ units (with $15$ hidden and $15$ visible variables). We derive the ground truth (GT) MAP estimates with brute force enumeration. \texttt{pgmpy}'s MPLP consistently gives poor MAP estimates. \texttt{pomegranate} recovers the GT MAP estimates for only $2/20$ RBMs, but often gives high-quality MAP estimates. PGMax recovers the GT MAP estimates for $17/20$ RBMs, and outperforms both \texttt{pgmpy} and \texttt{pomegranate} for each RBM. In addition, to compare inference time, we run inference $10$ times for each of the $20$ RBMs and report the average time. When compared with PGMax on CPU, \texttt{pgmpy} is $2.5\times$ slower, and \texttt{pomegranate} is $1.43\times$ slower. When compared with PGMax on GPU, \texttt{pgmpy} is $6\times$ slower, while \texttt{pomegranate} is $3.4\times$ slower.

We additionally test the scalability of each package to larger models, using randomly generated RBMs of increasing sizes---with $40, 60, \ldots, 200$ total units, and an equal number of hidden and visible variables. Again, \texttt{pgmpy}'s MPLP consistently gives poor MAP estimates. \texttt{pomegranate} and PGMax give similar-quality MAP estimates, with PGMax obtaining better MAP estimates for $8/9$ model sizes. However, as we see in Fig.~\ref{fig:results}, PGMax achieves significant inference speedups. This speed advantage is more pronounced for large models: for a RBM with $200$ units, inference takes $2.31$s with PGMax on a CPU, $0.84$s with PGMax on a GPU, $377.97$s with \texttt{pgmpy}'s MPLP, and $941.87$s with \texttt{pomegranate}. Note that our tests are done using a laptop GPU: the performance gap would increase for larger models on more powerful GPUs.

\subsection{Specifying complex factor graphs}\label{sec:fg}

In \href{https://github.com/deepmind/PGMax/blob/main/examples/rcn.ipynb}{one of our example notebooks}, we demonstrate a PGMax implementation of max-product LBP inference for Recursive Cortical Network (RCN)~\citep{george2017generative}. By explicitly enumerating the valid configurations of a factor, PGMax naturally supports RCN's non-standard pairwise factors. 
\href{https://github.com/deepmind/PGMax/blob/main/examples/pmp_binary_deconvolution.ipynb}{Another example notebook} uses PGMax to specify a PGM with logical factors, and leverages the message updates with linear complexity to efficiently solve the 2D blind deconvolution problem from \citep{lazaro2021perturb}.
Finally, PGMax's fully flat array-based LBP implementation allows it to support (a) factor graphs with arbitrary topology; (b) different types of factors and (c) discrete variables with a varying number of states, without compromises on speed or memory usage---which existing alternatives struggle to do.

\subsection{Exciting new possibilities from implementing LBP in JAX}\label{sec:jax}

PGMax implements LBP as pure functions with no side effects. This functional design means that we can easily apply JAX transformations like \texttt{jit/vmap/grad}, etc., to these functions, and additionally allows PGMax to seamlessly interact with other packages in the rapidly growing JAX ecosystem. This opens up some exciting new possibilities, like processing batches of samples/models in parallel using \texttt{vmap}, and allows to use of LBP as part of a larger end-to-end differentiable system.

\bibliographystyle{unsrtnat}
\nobibliography*
\bibliography{refs}

\end{document}